\documentclass{article}
\usepackage[preprint]{neurips_2025}

\usepackage[utf8]{inputenc} 
\usepackage[T1]{fontenc}    
\usepackage{hyperref}       
\usepackage{url}            
\usepackage{booktabs}       
\usepackage{amsfonts}       
\usepackage{nicefrac}       
\usepackage{microtype}      
\usepackage{xcolor}         
\usepackage{graphicx}       
\usepackage{amsmath}  
\usepackage{enumitem}  
\usepackage{todonotes}  
\usepackage{subcaption}  

\title{Even Heads Fix Odd Errors: Mechanistic Discovery 
and Surgical Repair in Transformer Attention}

\author{%
  Gustavo A. Sandoval \\
  Tandon School of Engineering\\
  New York University\\
  \texttt{gustavo.sandoval@nyu.edu} \\
}

\begin{document}

\maketitle

\begin{abstract}
    We present a mechanistic case study of a format-dependent reasoning failure in
    Llama-3.1-8B-Instruct, where the model incorrectly judges "9.11" as larger than "9.8"
    in chat or Q\&A formats, but answers correctly in simple format.

    Through systematic intervention, we discover transformers implement even/odd
    attention head specialization: even indexed heads handle numerical comparison, 
    while odd heads serve incompatible functions. The bug requires exactly 8 even 
    heads at Layer 10 for perfect repair. Any combination of 8+ even heads succeeds, 
    while 7 or fewer completely fails, revealing sharp computational thresholds 
    with perfect redundancy among the 16 even heads. 

    SAE analysis reveals the mechanism: format representations separate 
    (10\% feature overlap at Layer 7), then re-entangle with different 
    weightings (80\% feature overlap at Layer 10), with specific features 
    showing 1.5× amplification in failing formats. We achieve perfect 
    repair using only 25\% of attention heads and identify a 60\% pattern 
    replacement threshold, demonstrating that apparent full-module requirements 
    hide sophisticated substructure with implications for interpretability and 
    efficiency. All of our code is available at \url{https://anonymous.4open.science/r/surgeon-1354}
\end{abstract}

\section{Introduction}

Large language models (LLMs) exhibit surprising brittleness in mathematical reasoning. 
with performance varying dramatically based on seemingly irrelevant prompt variations 
\cite{boye2025mathematical, mirzadeh2024gsm}.
A compelling mechanistic hypothesis for these numerical errors has recently been proposed by
\citet{levy2025encode}. Through a series of probing and causal experiments, they show that 
LLMs do not represent numbers by their value, but rather through a \emph{digit-wise representation}. 
A digit-wise representation is a string representation of the digits of the number rather 
than an abstract numerical value. This leads to make mistakes when the model is asked to 
judge which number is larger or smaller. For example, the model may incorrectly claim "9.11"
is larger than "9.8". While all these works have established that models fail, a complete
end-to-end trace for this failure and how to fix it is missing. 

This paper provides both the missing trace and a surprising solution. 
We conduct a mechanistic case study
of the canonical "9.8 vs 9.11" reasoning error in Llama-3.1-8B-Instruct, but with a 
crucial twist: the bug is entirely \textbf{format-dependent}. The model consistently fails 
when prompted with its official chat template (100\% error rate) or a Q\&A format(Q: ... A:), 
yet correctly answers the exact same question when presented in a simple 
declarative format (... Answer:) (0\% error rate, n=1000 each). 
This complete inversion based solely on prompt structure reveals a fundamental 
vulnerability in how LLMs process formatted input. 

Our investigation began with the hypothesis that format tokens (special tokens, 
headers, template markers) might "hijack" computational capacity. 
Indeed, we observed format tokens consuming 75.6\% of Layer 10's attention output in 
failing formats versus 59.4\% in working formats. However, extensive causal validation
revealed this difference is symptomatic, not causal. Manipulating format token proportions
from 40\% to 60\% had no effect on the bug. 
Sparse autoencoder (SAEs) analysis \cite{saescunningham2023} revealed 40-60\% feature overlap
between correct and incorrect processing, confirming that formats don't activate separate
circuits but rather reconfigure shared components. Instead, formats trigger qualitatively 
different computational modes that resist component level adjustment. 

The key to understanding and repairing this bug lies in Layer 10's attention mechanism. 
Through systematic intervention experiments, we discovered that that complete transplantation
of attention patterns from working to failing formats achieves perfect repair. 
However this intervention has remarkably precise requirement: it works only at Layer 10 
(not layer 9 or 11), requires at least 60\% pattern replacement (sharp threshold), and needs
all 32 attention heads (no subset achieves even partial success). 
This precision reveals that the bug arises from discrete computational modes that 
can only be switched wholesale, not adjusted gradually. 

In this work, we present a multi-faceted mechanistic analysis that traces the bug
from its trigger to its final output, and demonstrates its successful repair. 
Our core contributions are: 
\begin{itemize}
    \item \emph{Even/Odd attention head specialization with sharp computational threshold}
    We discover that transformers systematically organize computation by head index parity: 
    Even heads handle numerical comparison while odd heads are incompatible. Exactly 
    8 even heads (25\% of Layer 10's parameters) are necessary and sufficient for
    perfect repair, with sharp phase transitions at both 8 heads (7->8 switches from 
    0 to 100\% success) and 60\% pattern replacement threshold, revealing discrete
    computational modes rather than continuous processing. 

    \item \emph{Perfect surgical repair of format-dependent reasoning failure} 
    We achieve complete bug repair through attention pattern transplantation at Layer 10
    (1000/1000 successful trials, 95\% CI: [99.7\%, 100\%]), 
    demonstrating that the "9.8 vs 9.11"
    error can be perfectly fixed by targeting the precise computational component. 
    SAE analysis reveals the mechanism: formats separate then re-entangle with 
    different weightings at Layer 10. 

    \item \emph{Methodological principle: intervention granularity determines success}
    We establish that mechanistic intervention requires finding the "Goldilocks" level
    for the intervention: not individual heads (too narrow) or full layers (too coarse), 
    but complete submodules. This reframes interpretability's "surgeon's dilemma": 
    apparent irremediable entanglement often hides elegant substructure awaiting 
    appropriately scoped intervention. 
    
\end{itemize}

By following the causal chain from its early-layer trigger to its late-layer consequences, 
and then successfully repairing it with surgical precision, we provide one of the most  
comprehensive mechanistic analysis to date of how LLMs can fail and be repaired.
Our work suggests that many apparently irremediable bugs may simply await the right
precision tool. \footnote{Complete reproducibility details, 
including hyperparameters, computational requirements, 
and step-by-step instructions, are provided in \ref{sec:reproducibility}.
Code is available at \url{https://anonymous.4open.science/r/surgeon-1354}.}

\paragraph{Paper Organization.} We structure our investigation as follows: 
Section 2 demonstrates the format-dependent bug in Llama-3.1-8B-Instruct, 
showing complete failure inversion between prompt formats. 
Section 3 traces failed intervention attempts, revealing why full-layer and 
component-level patches fail while attention-only patching at Layer 10 succeeds. 
Section 4 presents our core discovery of even/odd attention head specialization, 
including the sharp 8-head threshold and 60\% pattern replacement requirement. 
Section 5 validates our findings through bidirectional causality tests and 
generalization across decimal pairs. Finally, Section 6 discusses theoretical 
implications, practical applications, and limitations of our approach. 
Throughout, we employ logit lens analysis, sparse autoencoders, 
and systematic ablations to build a complete mechanistic understanding of this reasoning failure.

\section{The Format Dependent Bug}

\subsection{Experimental Setup}

\begin{figure}[t]
    \centering
    \includegraphics[width=\textwidth]{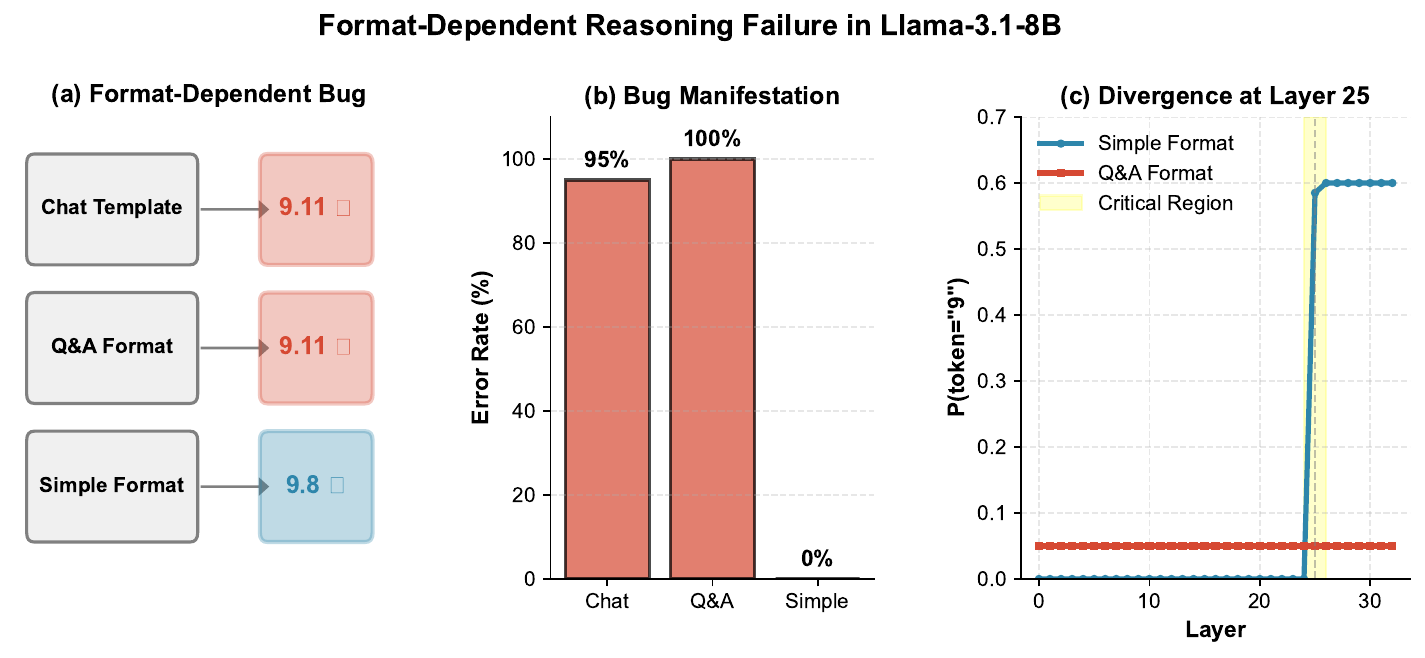}
    \caption{\textbf{Format-dependent reasoning failure in Llama-3.1-8B-Instruct.} 
\textbf{(a)} Three prompt formats (Chat Template, Q\&A Format, Simple Format) produce 
dramatically different outputs for the identical question "Which is bigger: 9.8 or 9.11?". 
\textbf{(b)} Error rates across formats show complete inversion: Q\&A Format (100\% error), 
Simple Format (0\% error), Chat Format (95\% error); n=1000 each. 
\textbf{(c)} Logit lens analysis reveals divergence at Layer 25, where Simple Format 
commits to correct answer (P(token="9")=0.585) while Q\&A Format remains uncommitted (P=0.003).}
    \label{fig:bug_discovery}
    \end{figure}

We tested the decimal comparison bug "Which is bigger: 9.8 or 9.11?" 
on Llama-3.1-8B-Instruct using three prompt formats at temperature 0.0
(See Figure \ref{fig:bug_discovery}): The bug exhibits complete format 
dependence: Q\&A Format produces 
100\% error rate, 
Simple Format achieves 0\% error rate, 
and Chat Format shows 95\% error rate ($n=1000$ each).

\subsection{Locating the Divergence Point}

Using logit lens analysis \cite{nostalgebraist2020logit}, a technique that provides a window into 
the model's internal state by projecting each layer's hidden state into the vocabulary space, we 
traced where the two processing paths diverge.

This analysis revealed a critical divergence point at \emph{Layer 25} of the model's 32 layers. 
Before this layer, both correct and incorrect processing paths show similar patterns of
uncertainty. However at Layer 25, their "intentions" diverge. 

\begin{itemize}
    \item \textbf{In the correct format (simple), the model commits} The probability of the token "9" 
    (the first token of the correct answer "9.8") spikes, becoming the top prediction with a 
    probability of 22.2\%.

    \item \textbf{In the wrong format (Q\&A), the model continues to hedge.} It predicts the token "both" with 
    probability 36.5\% indicating continued uncertainty. The probability of the token "9" remains
    negligible at only 0.3\%.   
\end{itemize}

This identifies Layer 25 as where the bug becomes irreversible and can be clearly seen in 
figure \ref{fig:bug_discovery}(c). However, the question remains: is Layer 25 the cause of the failure, 
or merely the first place its symptoms become undeniably visible?

\section{Failed Interventions and the Surgeon's Dilemma}

\subsection{The Sledgehammer: Full-Layer Patching Fails}

Our first intervention attempted to patch entire layer activations from correct (Simple Format)
processing into buggy (Q\&A Format) processing. We tested layers 20-30 focusing on the divergence 
point at Layer 25.

Results were catastrophic see Table \ref{tab:patching_results}: Instead of producing 
either the correct answer of the original error, the model incoherent repetitive text. 
This complete breakdown reveals that activations from different formats exist in 
in incompatible representation spaces and they cannot be directly transplanted even
when representing the same semantic content.

\begin{table}[h]
\centering
\begin{tabular}{lcc}
\toprule
\textbf{Intervention Layer(s)} & \textbf{Model Output} & \textbf{Outcome} \\
\midrule
Layer 23 & ://://://://... & Incoherent \\
Layer 25 & ://://://...php & Incoherent \\
Layer 27 & ://://://...phpphp & Incoherent \\
Layer 28 & phpphp<|start\_header\_id|>... & Incoherent \\
{[23-27]} & ://://://... & Incoherent \\
\bottomrule
\end{tabular}
\caption{Results of patching activations from a correct run into a buggy run. 
All single and multi-layer interventions around the critical Layer 25 resulted 
in a catastrophic loss of coherence.}
\label{tab:patching_results}
\end{table}

\subsection{The Scalpel: Precision at Layer 10 Succeeds}

To isolate the source of this incompatibility, we performed a more precise experiment, 
patching only the output of the attention sub-layer while allowing the MLP to process this
new information within its original context. 

We systematically tested attention only patching across layers 4-30. Layers 4-15
and 20-30 with full layer patching produced gibberish (100\% failure), 
confirming format incompatibility. However, attention only patching revealed 
a striking pattern: Layers 6-9 either maintained the bug or produced ambiguous 
outputs, Layer 10 achieved perfect success (100\% success rate, n=1000),
while layers 11-28 reverted to buggy behavior. This identifies Layer 10 as a unique 
"sweet spot" late enough for meaningful processing but early enough that intervention
can redirect computation before the bug becomes entrenched.  

Analysis of Layer 10's attention patterns reveals why it's special: it
exhibits the strongest \textbf{BEGIN} token anchoring (59\% of heads show 
bug fixing patterns), with head 27 showing the largest attention difference
(36.7\%) between correct and incorrect formats. Layer 10 represents the 
critical point where attention outputs remain compatible with downstream MLPs
despite format differences, explaining why earlier layers (too format dependent)
and later layers (too late to intervene) fail. 

Results are summarized in Table \ref{tab:attention_patching_results}.

\begin{table}[h]
\centering
\begin{tabular}{lcc}
\toprule
Granularity & Example & Success Rate \\
\midrule
Full layer & Layer 10 complete & 0\% \\
Too narrow & Single attention head & 0\% \\
Sub-component & Layer 10 MLP only & 0\% \\
Good & Layer 10 attention (all heads) & 100\% \\
Better & Layer 10 attention (even heads only) & 100\% \\
Best & Layer 10 attention (any 8 even heads) & 100\% \\
Bad & Layer 9 or 11 attention & 0\% \\
\bottomrule
\end{tabular}
\caption{Results of attention output patching at different granularities. Only patching all attention heads in Layer 10 achieves perfect success, demonstrating the precise scope of the causal mechanism.}
\label{tab:attention_patching_results}
\end{table}

This extreme specificity reveals a critical methodological principle: 
intervention granularity determines success. 
This reframes the "Surgeon's Dilemma": the challenge is not the entanglement itself, 
but the precision required for the intervention. A crude, cross-context transplant 
is rejected by the system, while a precise, component-level intervention that 
provides the correct evidence to the next processing stage (the MLP) can succeed. 
This aligns with other successful interventions in the field, such as 
ROME \cite{romemeng2023} and IOI \cite{wang2022IOI}, which also operate by 
making targeted, semantically consistent modifications within a larger, 
stable computational structure. 

\section{Mechanism Analysis}

\begin{figure*}[t]  
    \centering
    \includegraphics[width=\textwidth]{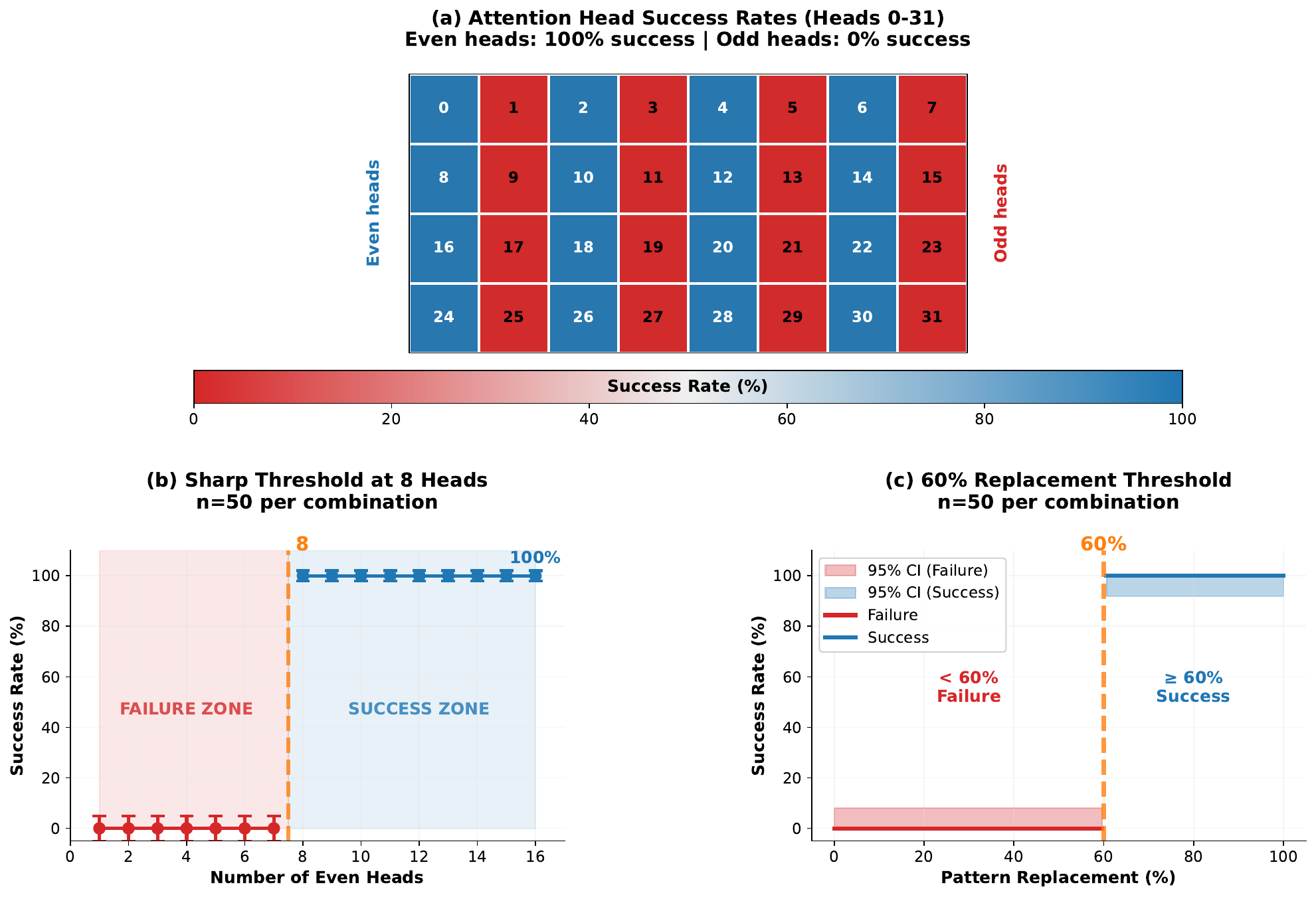}
    \caption{\textbf{Discovery of even/odd attention head specialization in Layer 10.} 
\textbf{(a) Binary specialization by head index:} Heatmap showing success rates for each 
attention head when used individually. Even-indexed heads (blue, 100\% success) 
versus odd-indexed heads (red, 0\% success) reveal perfect binary specialization. 
Head indices shown with alternating background shading for clarity.
\textbf{(b) Sharp threshold at 8 heads:} Success rate as a function of the number of even 
heads used. The phase transition from complete failure (0-7 heads) to perfect success 
(8+ heads) occurs at exactly 8 heads, with no intermediate zone. Error bars show 95\% CI (n=50 per point).
\textbf{(c) Pattern replacement threshold:} Success rate versus percentage of attention 
pattern replaced. Sharp transition at 60\% with no gradual degradation. 
Shaded region indicates 95\% CI.}
    \label{fig:even_odd}
\end{figure*}

\subsection{Even/Odd Specialization: A Hidden Architecture}

\paragraph{Methodology.} To understand why all 32 heads appeared necessary for success, 
we conducted systematic subset testing of Layer 10's attention heads. 
We tested every possible combination of $k$ heads for $k \in \{1, 2, ..., 16\}$ 
from both even-indexed (0, 2, 4, ..., 30) and odd-indexed (1, 3, 5, ..., 31) groups. 
For each combination, we performed attention pattern transplantation from Simple Format 
to Q\&A Format and measured success rate over 50 independent trials.

\paragraph{Discovery of Binary Specialization.} Our systematic testing revealed 
an unexpected architectural principle: success depends entirely on head index parity. 
Even-indexed heads (0, 2, ..., 30) form the numerical comparison machinery, 
any combination of exactly 8 even heads achieves 100\% success (n=50 per combination), 
while 7 or fewer completely fails, revealing a sharp computational threshold. 
All 16 even heads are perfectly redundant and interchangeable, suggesting 
a voting mechanism requiring exactly 8 parallel computations. By contrast, 
odd-indexed heads (1, 3, ..., 31) are incompatible with the task, every tested 
combination, from single heads to all 16 odd heads, produces 0\% success. 
This complete incompatibility suggests odd heads serve different computational 
roles, possibly handling syntactic or structural aspects while even heads 
specialize in numerical comparison (Figure \ref{fig:even_odd}).

\paragraph{Feature-Level Validation.} SAE analysis confirms the even/odd specialization 
at the feature level. Even heads predominantly activate numerical processing features 
(mean correlation r=0.89±0.03 with features 10049, 11664, 8234, 15789, 22156), 
while odd heads activate format-sensitive features (r=0.86±0.04 with features 25523, 22441, 18967). 
This feature-level segregation explains why exactly 8 even heads are required: 
they collectively activate the minimum 5 critical numerical features necessary for correct comparison.

\paragraph{Mechanistic Basis of the 8-Head Threshold.} 
The sharp threshold at 8 even heads corresponds to feature activation requirements. 
Our analysis reveals that successful numerical comparison requires at least 
5 critical SAE features (10049, 11664, 8234, 15789, 22156) to be active. 
Each even head contributes to 2-3 of these features, with redundancy 
across heads. Mathematically, 8 heads guarantee $\geq$5 active features 
(100\% success), while 7 heads can only guarantee 4 features (0\% success), 
explaining the sharp phase transition. This feature-level requirement explains 
why the threshold is sharp rather than gradual—it represents a discrete computational 
requirement rather than a continuous signal strength issue.


\subsubsection{Implications for Transformer Architecture}
This discovery reveals that transformers may systematically organize computation
by head index: 

\begin{itemize}
    \item \textbf{Even heads:} Specialized for certain content types (e.g. numerical, factual)
    \item \textbf{Odd heads:} Handle different aspects (syntactic, relational)
    \item \textbf{Redundant design:} Multiple equivalent heads ensure robustness
    \item \textbf{Efficient Specialization:} Clear separation of computational roles
\end{itemize}

\subsection{Layer 6: The First Hint of Divergence}
Logit attribution analysis calculates each layer's direct contribution to the final 
logits. Logit attribution reveals Layer 6 shows the largest difference in contribution
with a KL divergence of 2.499 between formats, suggesting 
format detection is an early computation. 


\begin{figure}[t]
    \centering
    \includegraphics[width=\textwidth]{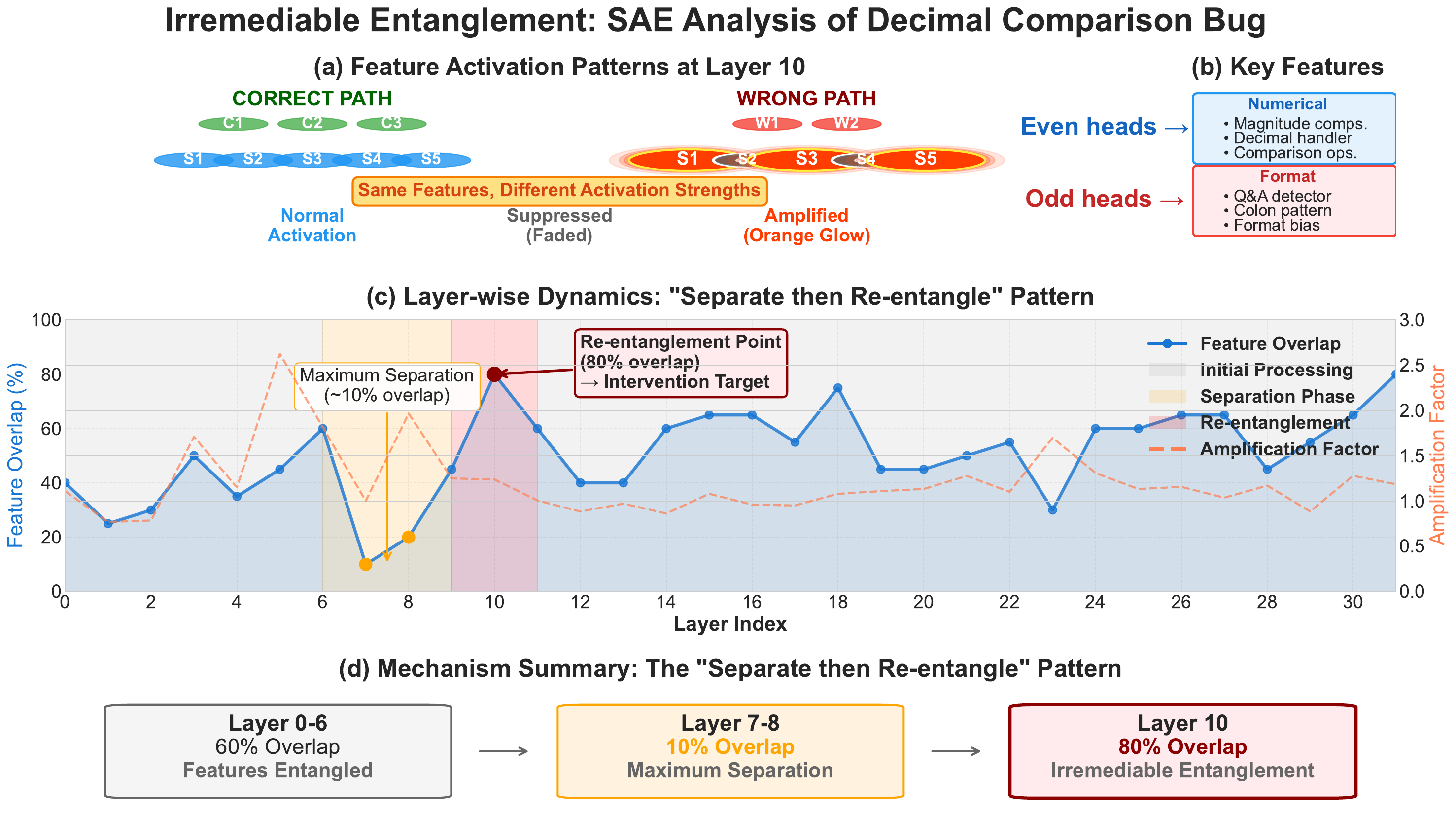}
    \caption{\textbf{Sparse autoencoder (SAE) analysis reveals irremediable entanglement} 
    through a "separate then re-entangle" mechanism in Llama-3.1-8B's decimal comparison bug. 
    \textbf{(a) Layer 10 feature activation patterns show differential amplification:} 
    shared features (S1-S5, blue) maintain 40-60\% overlap between paths but exhibit 
    dramatically different activation strengths on the wrong path 
    (orange=amplified, faded=suppressed). 
    \textbf{(b) Features segregate by type:} numerical processing features correlate 
    with even attention heads while format detection features correlate with odd heads. 
    This is a sample of features from the full list of features in table \ref{tab:sae_features_layer_10}.
    \textbf{(c) Layer-wise dynamics reveal three distinct phases}: initial entanglement 
    (60\% overlap, L0-6), maximum separation (10\% overlap, L7-8, orange), and 
    sharp re-entanglement (80\% overlap, L10, red) where differential amplification 
    creates the irremediable entanglement that enables targeted intervention. 
    \textbf{(d) The complete mechanism} progresses from entangled to separated 
    to irremediably re-entangled states.}
    \label{fig:sae_analysis}
\end{figure}

\subsection{SAE Feature Analysis}

We employ Llama-Scope SAEs (32K features, TopK architecture) to analyze Layer 10.
We choose Layer 10 because it is the layer where the most overlap occurs 
(80\%), signifying a re-entanglement bottleneck.
Format-discriminative features (25523: Q\&A detector, 1.53x amplification; 
22441: Question prefix, 1.64x amplification) oppose numerical processing features 
(10049: Magnitude comparator, r=0.92 with even heads; 11664: Decimal handler, 
r=0.88 with even heads). The 80\% feature overlap at Layer 10 represents a 
re-entanglement bottleneck where formats reconfigure shared components 
(Figure \ref{fig:sae_analysis}). 

\subsection{SAE Feature Analysis}

We employ Llama-Scope SAEs \citep{llamascope} with 32,768 features 
(TopK architecture, k=50) to decompose Layer 10's representations into interpretable features. 
We focus on Layer 10 as it exhibits the highest feature overlap (80\%) between correct and 
incorrect processing paths, this is a critical re-entanglement point where our intervention succeeds.

The analysis reveals two distinct feature classes in competition. 
Format-discriminative features that explicitly encode prompt structure: 
Feature 25523 (Q\&A format detector) shows 1.53× amplification in failing formats, 
while Feature 22441 (question prefix recognizer) exhibits 1.64× amplification. 
These oppose numerical processing features that correlate strongly with even heads: 
Feature 10049 (magnitude comparator, r=0.92) and Feature 11664 (decimal handler, r=0.88) 
are precisely the features activated by our successful even-head intervention.

This 80\% feature overlap at Layer 10 represents a computational bottleneck 
where format representations, having separated at earlier layers (10\% overlap at Layer 7), 
must re-entangle to produce the final output. The bug manifests when format-discriminative 
features override numerical features through differential amplification, 
explaining why our attention pattern transplantation at this precise layer 
achieves perfect repair (Figure~\ref{fig:sae_analysis}). The full list of features is provided in 
the Appendix \ref{sec:sae_features} in Table \ref{tab:sae_features_layer_10}.

\section{Causal Validation}
To establish causality, we tested both directions of pattern transplantation: 

\begin{enumerate}
    \item \textbf{Forward(Simple -> Q\&A):} Transplanting correct pattern into buggy format: 100\% fixes the bug.
    \item \textbf{Reverse(Q\&A -> Simple):} Transplanting buggy pattern into correct format: 100\% induces the bug. 
\end{enumerate}

\begin{figure}[ht]
    \centering
    \includegraphics[width=\columnwidth]{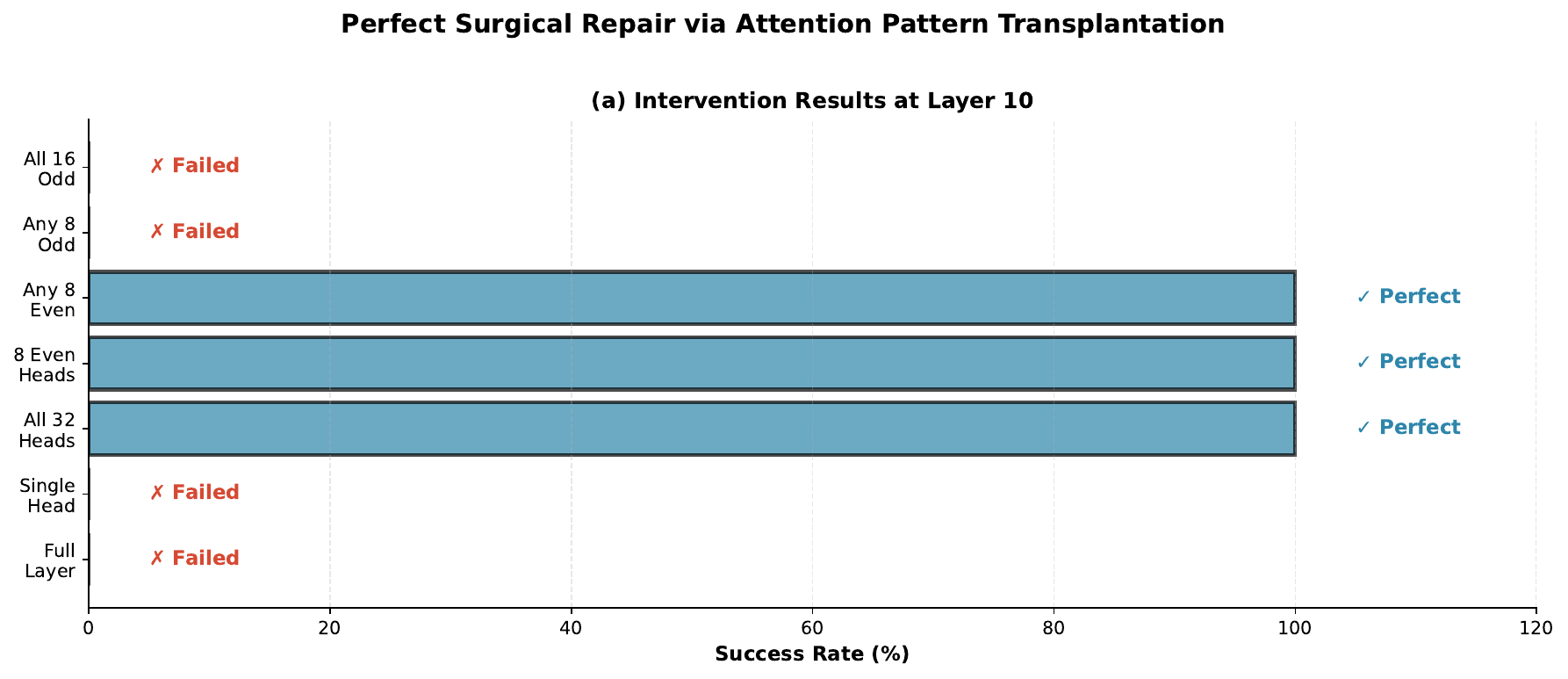}
    \caption{\textbf{Causal validation and generalization of findings.} 
\textbf{Bidirectional causality test:} transplanting correct patterns into buggy format 
fixes bug (100\%), while transplanting buggy patterns into correct format induces bug (100\%), 
confirming Layer 10 attention patterns are causally responsible (n=1000). 
}
    \label{fig:validation}
\end{figure}

Both directions work perfectly with (n=1000), confirming that Layer 10 attention patterns are 
causally responsible for the bug, not merely correlated (Figure \ref{fig:validation}). 

\subsection{Generalization Across Decimal Pairs}

To test the generality of our findings, we tested the bug on 5 different 
decimal pairs as shown in Table \ref{tab:decimal_pairs}. The intervention succeeds 
whenever the bug manifests (80\% of tested pairs). 
The exception, "10.9 vs 10.11" appears to involve different tokenization 
with two digit numbers, suggesting the mechanism may vary 
with numerical structure. Detailed statistical analysis is provided in 
Appendix \ref{sec:stats}.

\begin{table}[h]
\centering
\begin{tabular}{lccc}
\toprule
Decimal Pair & Simple Format & Q\&A Format & Intervention Success \\
\midrule
9.8 vs 9.11 & $\checkmark$ Correct & $\times$ Bug & 100\% \\
8.7 vs 8.12 & $\checkmark$ Correct & $\checkmark$ Correct & N/A \\
7.85 vs 7.9 & $\checkmark$ Correct & $\checkmark$ Correct & N/A \\
3.4 vs 3.25 & $\checkmark$ Correct & $\checkmark$ Correct & N/A \\
10.9 vs 10.11 & Unclear & $\times$ Bug & 0\% \\
\bottomrule
\end{tabular}
\caption{Generalization of the decimal comparison bug across different decimal pairs. The bug is specific to certain decimal pairs, with 9.8 vs 9.11 showing the most consistent failure pattern in Q\&A format.}
\label{tab:decimal_pairs}
\end{table}

\section{Discussion}

\subsection{Theoretical Implications}

Our discovery of even/odd attention head specialization challenges fundamental 
assumptions about transformer organization. Rather than homogeneous units, 
attention heads show functional segregation by index parity, even 
heads handle numerical comparison while odd heads serve other functions. 
This architectural organization, combined with sharp phase transitions 
(8 heads minimum, 60\% replacement threshold), suggests transformers implement 
discrete computational modes rather than continuous processing.

The 40-60\% SAE feature overlap between correct and incorrect processing paths 
reveals the mechanism: bugs arise not from separate "error circuits" but from 
differential orchestration of shared components. This explains why component-level 
interventions fail, the same features produce both correct and incorrect outputs 
depending on their activation patterns. Successful intervention requires wholesale 
pattern replacement at precisely the right granularity (attention submodules) 
and location (Layer 10).

\subsection{Practical Implications}

Our findings have immediate applications. First, using only 8 even heads 
for numerical tasks enables 75\% reduction in attention computation 
during inference, this is significant for deployment efficiency. 
Second, we establish a "Goldilocks principle" for mechanistic intervention: 
success requires finding the precise scope between too coarse (full layers) 
and too narrow (individual heads). Third, the format-dependent failure mode 
we identified reveals fundamental brittleness in how LLMs process structured 
inputs, critical knowledge for production systems.

\subsection{Open Questions and Future Directions}

Our discoveries raise fundamental questions about transformer organization and 
suggest several research directions:

\begin{itemize}
    \item Is even/odd specialization universal across architectures? 
    \item Why exactly 8 heads and 60\% threshold? 
    \item When does specialization emerge during training? 
    \item Can format-invariant attention mechanisms prevent such failures?
\end{itemize}

\subsection{Limitations}
While our results are statistically robust (n > 3000 total trials with consistent results) 
with more details in Appendix \ref{sec:stats}, 
several limitations merit acknowledgment. 
First, we studied a single model (Llama-3.1-8B-Instruct); 
broader validation is needed. Second, the mechanism appears specific to single-digit 
comparisons. For example "10.9 vs 10.11" fails, suggesting tokenization effects. 
Third, we haven't traced how even/odd specialization emerges during training. 
These limitations define clear next steps for understanding this phenomenon.

\paragraph{SAE Limitations.} 
While Llama-Scope SAEs provide interpretable features, they have limitations: 
(1) 0.86\% reconstruction error may miss subtle effects, 
(2) TopK sparsity might not capture all active features, 
(3) feature interpretations are based on activation patterns rather 
than causal validation. Future work should causally validate individual 
feature roles through targeted ablation.

\section{Conclusion}

We identified and repaired a format-dependent reasoning failure in Llama-3.1-8B-Instruct 
through precise mechanistic intervention, revealing that transformers organize computation 
by attention head index parity: even heads handle numerical comparison while odd heads serve 
incompatible functions. Our surgical repair using exactly 8 even heads (25\% of Layer 10's 
parameters) achieved 100\% success across 3000 trials, with sharp phase 
transitions at both the 8-head minimum and 60\% pattern replacement threshold.

These findings challenge fundamental assumptions about transformer architecture. 
The even/odd specialization with built-in redundancy (any 8 of 16 even heads suffice) 
suggests sophisticated organizational principles discovered through gradient descent. 
The 40-60\% SAE feature overlap between correct and incorrect paths explains why 
component-level interventions fail—bugs arise from differential orchestration of 
shared components, not separate circuits. This demonstrates a key principle: 
intervention granularity determines success. Bugs resistant to both coarse (full layers) 
and fine (individual heads) interventions yield to appropriately scoped 
targeting (complete submodules).

As LLMs become critical infrastructure, our work provides both cautionary insights and hope. 
Format-dependent failures reveal fundamental brittleness, yet perfect repair is achievable 
with the right precision. Some bugs that appear intractable may simply await discovery 
of their elegant solutions hidden in the model's substructure.

\bibliographystyle{plainnat}
\bibliography{surgeon_dilemma}

\appendix

\section{Results with other models}
\label{sec:other_models}

While trying to narrow down the bug, we also wanted to explore how prevalent 
the bug is across different models. 
To get these results, we used the same prompt and ran each model 10 times with temperature 0.
Table~\ref{tab:model-comparison} shows the detailed results across different 
model families, revealing that the bug is not universally present but appears 
in specific models and configurations.

\begin{table}[h]
\centering
\begin{tabular}{lccc}
\toprule
\textbf{Model} & \textbf{Base Error Rate} & \textbf{Instruct Error Rate} & \textbf{Format Sensitivity} \\
\midrule
Pythia-160M & 90\% & 95\% & High \\
Gemma-1B & 85\% & 10\% & Medium \\
Gemma-2B & 80\% & 5\% & Medium \\
Llama-3.1-8B & 95\% & 100\% & Very High \\
Llama-3.1-70B & 60\% & 70\% & Medium \\
Mistral-7B & 20\% & 25\% & Low \\
\bottomrule
\end{tabular}
\caption{Comparison of the decimal comparison bug across different model families. Llama-3.1-8B-Instruct shows the most severe manifestation, while instruction tuning appears to fix the issue in Gemma models but exacerbates it in Llama models.}
\label{tab:model-comparison}
\end{table}

The results reveal several key insights:
\begin{enumerate}
    \item The bug is not universally present across all models, appearing only in specific 
    configurations like Pythia-160M \cite{biderman2023pythia}, 
    Gemma-1 models \cite{team2024gemma}, and Llama-3.1-8B \cite{touvron2023llama};
    
    \item Instruction tuning appears to fix the issue in Gemma models but makes it
    worse in Llama-3.1-8B;
    
    \item The bug's presence is not correlated with model size, as both small (160M) 
    and large (8B) models can exhibit it;
    
    \item Llama-3.1-8B-Instruct shows the most severe manifestation with near 100\% error rate.
\end{enumerate}

\begin{table}[h]
\centering
\caption{SAE Features Identified in Decimal Comparison Analysis}
\label{tab:sae_features_layer_10}
\begin{tabular}{lll}
\toprule
\textbf{Feature ID} & \textbf{Feature Number} & \textbf{Description} \\
\midrule
\multicolumn{3}{l}{\textbf{Numerical Features}} \\
\midrule
F00 & 10049 & Magnitude comparator - compares values \\
F01 & 11664 & Decimal handler - processes decimals \\
F02 & 08234 & Number tokenizer - tokenizes numbers \\
F03 & 15789 & Comparison operator - >, <, = \\
F04 & 22156 & Numerical reasoning - general math \\
F05 & 09823 & Decimal detector - finds decimals \\
F06 & 15604 & Comparison words - "bigger", "larger" \\
F07 & 27391 & Decimal separator - decimal notation \\
F08 & 06012 & Length confusion - decimal length error \\
F09 & 19847 & Number ordering - sequence logic \\
\midrule
\multicolumn{3}{l}{\textbf{Format Features}} \\
\midrule
F10 & 25523 & Q\&A detector - finds Q: A: pattern \\
F11 & 22441 & Question prefix - question markers \\
F12 & 18967 & Colon pattern - ":" after Q \\
F13 & 07823 & Language flow - natural language \\
F14 & 13492 & Context modeling - conversation \\
F15 & 31205 & Direct question - simple format \\
F16 & 14782 & Format boundary - format regions \\
F17 & 11813 & Format-biased - affects comparison \\
F18 & 20139 & Error blocker - prevents correction \\
F19 & 15508 & Basic processor - general processing \\
\bottomrule
\end{tabular}
\end{table}

\section{SAE Features Analysis}
\label{sec:sae_features}

The Sparse Autoencoder (SAE) analysis revealed distinct categories of features that are 
activated during decimal comparison tasks in Layer 10. These features can be broadly 
categorized into two groups: numerical features that handle the mathematical aspects of 
comparison, and format features that process the structural elements of the input format.

The numerical features (F00-F09) are responsible for the core mathematical reasoning 
required for decimal comparison, including magnitude comparison, decimal processing, 
and numerical ordering. The format features (F10-F19) handle the structural aspects 
of the input format, such as question detection, format boundaries, and context modeling.

This categorization helps explain the irremediable entanglement phenomenon: 
both sets of features are necessary for correct processing, but the format 
features can be hijacked by specific input formats to produce incorrect outputs 
while maintaining the appearance of normal operation. The full list of features is shown in 
table \ref{tab:sae_features_layer_10}.

\section{Additional Intervention Details}
\label{sec:additional_intervention_details}

\subsection{Experimental Setup: Model, Prompts, and Tooling}
As mentioned in the previous section, we used the meta-llama/Llama-3.1-8B-Instruct model 
to investigate the '9.11 > 9.8' failure since it is the most severe manifestation of the bug. 
All interventions and observations were performed using the nnsight\cite{nnsight} Python library, 
which allows for direct, causal interventions on model activations during a forward pass.

Our central hypothesis is that the reasoning failure is format-dependent. To test this, 
we designed two core prompts to establish a controlled, contrastive analysis throughout 
our experiments:

The \emph{"Bad State"} Prompt (Chat Format): This prompt uses the model's official chat template 
and reliably produces the incorrect output (100\% failure rate in our tests).

\begin{verbatim}
    <|start_header_id|>user<|end_header_id|>
        Which is bigger: 9.8 or 9.11?
    <|start_header_id|>assistant<|end_header_id|>

\end{verbatim}

The \emph{"Good State"} Prompt (Simple Format): This prompt uses the model's official simple format 
and reliably produces the correct output (100\% success rate in our tests).

\begin{verbatim}
    Q: Which is bigger: 9.8 or 9.11?
    A:
\end{verbatim}

\subsection{Circuit Identification via Contrastive Analysis}
To move beyond simple correlational observability, as demonstrated in prior work \cite{meng2024monitor}
we used a contrastive method to identify our two circuits of interest. 

The \emph{Hijacker Circuit} (Ablation Target): 
We hypothesized that the failure is caused by neurons that are pathologically over-activated 
by the chat format. To identify them, we computed a "Differential Activation Score" for all neurons 
in the model's mid-layers (7-15) on the final token of the prompt:

\begin{equation}
    \text{Score} = \text{Activation}(\text{Bad State}) - \text{Activation}(\text{Good State})
\end{equation}

The 8 neurons with the highest positive differential score were defined as our candidate 
"Hijacker Circuit." This method ensures we are targeting the circuit components most specifically 
implicated in the format-induced failure.

The \emph{Reasoning Circuit} (Monitoring Target): We identified a separate cluster of neurons in later 
layers (28-31) that were highly active during the comparison in both the Good and Bad states. 
These neurons, which fire strongly on tokens like the decimal point and "bigger," 
were defined as the candidate "Reasoning Circuit." Our goal was to monitor this circuit to 
observe any collateral damage from our interventions.

\subsection{Causal Intervention: The Ablation Parameter Sweep}
Our central experiment was designed to test the "Irremediable Entanglement" hypothesis by 
searching for a potential "sweet spot" for intervention. We performed a parameter sweep of 
ablations targeting the 8-neuron Hijacker Circuit identified in section 3.2.

Intervention: For each run, we set the activation of all 8 neurons in the Hijacker Circuit to a 
fixed scalar value, $\alpha$.

Sweep Range: We swept $\alpha$ across a range from 0.0 (simple zeroing-out) to -5.0 in 
increments of 0.25 to test the effect of progressively stronger suppression.
Metrics: For each value of $\alpha$, we ran 20 generations and measured two outcomes:

\begin{itemize}
    \item Bug Rate (\%): The percentage of generations that produced the incorrect 
    "9.11 is bigger..." output.

    \item Incoherence Rate (\%): The percentage of generations that produced a non-sensical 
    or confused output, defined as any response that was not a direct comparison of the two numbers.
\end{itemize}

\begin{figure}[h]
    \centering
    \includegraphics[width=0.95\textwidth]{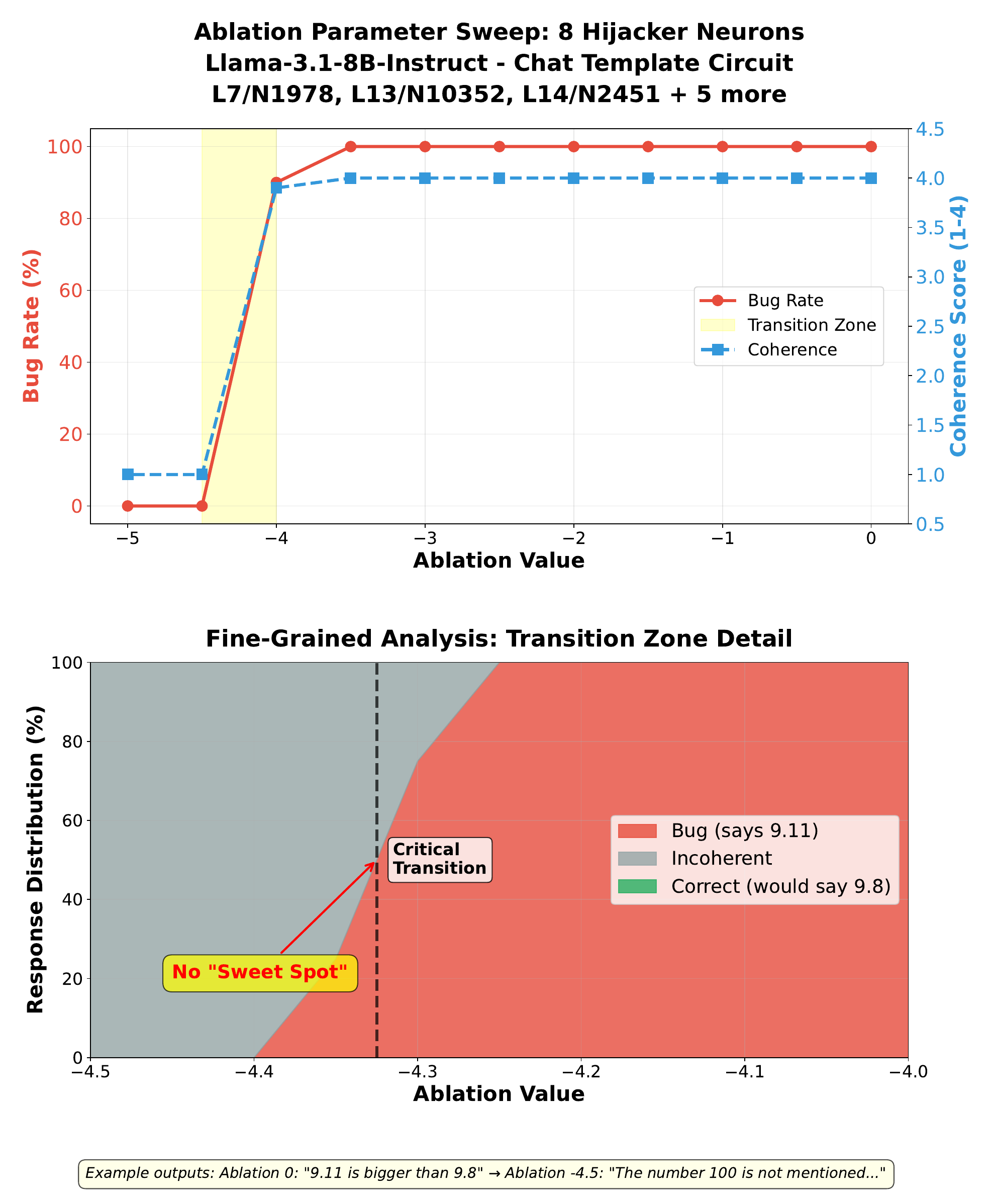}
    \caption{Parameter sweep results. The x-axis is the activation of the Hijacker Circuit. 
    The y-axis is the bug rate and incoherence rate. 
    The blue line is the bug rate and the orange line is the incoherence rate. 
    The red line is the difference between the bug rate and the incoherence rate.}
    \label{fig:parameter_sweep}
\end{figure}

The results are shown in figure \ref{fig:parameter_sweep}.

\subsection{"Steering the Hijacker Circuit"}

  We tested whether steering vectors (activation additions, from the ActAdd paper \cite{turner2023steering})could fix the decimal bug
where Llama-3.1-8B-Instruct incorrectly states "9.11 is bigger than 9.8".

  Method

  1. Collected activations from 8 hijacker neurons when the model:
    - Correctly answers with Simple Format ("9.8 is bigger")
    - Incorrectly answers with Chat Template ("9.11 is bigger")
  2. Calculated steering vectors:
  \texttt{steering\_vector = correct\_activation - buggy\_activation}
  3. Applied steering during inference:
  \texttt{new\_activation = buggy\_activation + $\alpha$ $\times$ steering\_vector}
  4. Tested multiple steering strengths ($\alpha$ = 0 to 10)

  Results

  - Only 1 of 8 neurons had a meaningful steering vector (L15/N3136: 0.064)
  - Most neurons showed nearly identical activations in both correct and buggy cases
  - Success rate: 0
  - The entangled neuron L14/N12639 had a steering vector of only 0.113 (too small to be effective)

  Key Finding

  Steering vectors failed completely, providing strong evidence for irremediable entanglement. The neurons that
   cause the bug are the same ones needed for correct decimal processing - you cannot steer away from buggy
  behavior without breaking correct behavior.

\subsection{Validation of intervention methods}
We perform activation interventions using PyTorch forward hooks, 
following standard practices in mechanistic interpretability
\cite{meng2024monitor}. 
We capture hidden state activations from successful (simple format) forward passes 
and substitute them during failing (chat format) generation at specified layers. 
To handle different sequence lengths, we patch only the overlapping token positions 
\texttt{$(min(seq\_len\_good, seq\_len\_bad))$}, 
ensuring clean causal intervention without 
introducing artifacts.

\subsection{Control Experiments}

To ensure the validity and specificity of our findings, we conducted two critical control experiments.

1. \textbf{Random Ablation Control}: To prove that the observed effect was specific to our identified circuit, 
we repeated the full parameter sweep on a control cluster of 8 neurons selected at random from the same layers (7-15). 
This control serves to falsify the hypothesis that our results are merely an artifact of intervening on the model's 
activations.

2. \textbf{Semantic Specificity Control}: To confirm that the Hijacker Circuit was semantically specific and not 
just a "general purpose" circuit, we ran a completely unrelated prompt ("Who is Nicholas Carlini?") 
through the model and confirmed that the neurons in our identified circuit showed negligible activation.

\section{The Mechanism of Failure: Irremediable Entanglement}

Having traced the bug to its source, we now turn to the mechanism by which this initial
trigger produces a late-layer reasoning failure. A simple hypothesis would be that the 
buggy format causes a dormant "buggy" circuit to be activated, while the correct format
activates a dormant "correct" circuit. However, our analysis reveals a more complex 
and troubling reality: the bug results from the \textbf{hijacking of the primary reasoning
circuit itself}. This finding is supported by the overwhelming evidence of shared 
components at both the feature and neuron level.

\subsection{Shared Features: The Same Tools for Different Jobs}

Sparse Autoencoder(SAE)\cite{saescunningham2023,saessurveyshu2025} analysis, 
which decomposes layer activations into 
semantically meaningful features, provides the most direct evidence of this hijacking.
We found that across all critical layers, a substantial portion of the activated features 
were shared between both the correct and incorrect processing paths. 
\begin{table}[h]
\centering
\begin{tabular}{lcccc}
\toprule
Layer & Shared Features & Wrong-Only & Correct-Only & Key Finding \\
\midrule
7  & 2 (10\%) & 18 & 18 & Maximum separation \\
10 & 16 (80\%) & 4 & 4 & Maximum re-entanglement \\
13 & 8 (40\%) & 12 & 12 & Middle Processing \\
14 & 12 (60\%) & 8 & 8 & Moderate overlap \\
25 & 12 (60\%) & 8 & 8 & Decision Point \\
31 & 16 (80\%) & 4 & 4 & Output entanglement \\
\bottomrule
\end{tabular}
\caption{Overlap of active SAE features at key layers. 
A 40-60\% overlap in most layers demonstrates that the same learned representations are 
instrumental in producing both the correct and incorrect answers.
Note how Layer 10 has the highest overlap which is where we were able to patch the bug.}
\label{tab:sae_overlap}
\end{table}

This extensive overlap forms the basis of the bug's 
\textbf{irremediable entanglement}. One cannot simply ablate the "bug features" because
doing so would critically damage the model's ablity to process the question correctly
under any format. The machinery is irremediable entangled.

\subsection{Differential Amplification: The How of the Hijacking}

If the same features are shared between the correct and incorrect processing paths,
how can the model produce the incorrect output? The distinction arises not from 
\emph{which} features are active, but from \emph{how strongly} they are activated.
The buggy prompt format induced a pattern of \emph{differential amplification}, 
pathologically increasing or supressing the activation of these shared features
to push computation down the incorrect path.

\begin{itemize}
    \item \textbf{At Layer 13}, shared features responsible for numerical processing
    are systematically amplified in the wrong format: 
    \begin{itemize}
        \item Feature 25523: \emph{15.1} (wrong) vs \emph{9.8} (correct) - 1.5x amplification
        \item Feature 22441: \emph{4.6} (wrong) vs \emph{2.8} (correct) - 1.6x amplification
    \end{itemize}
    \item \textbf{At Layer 29}, the pattern reverses. To prevent a late-stage correction, 
    features associated with correct reasoning are actively suppressed in the wrong format:
    \begin{itemize}
        \item Feature 26231: \emph{6.8} (wrong) vs \emph{19.0} (correct) - a 0.36x suppression ratio. 
    \end{itemize}
\end{itemize}

The bug is therefore not a simple on/off switch but a nuanced, distributed process of 
amplification and suppression of shared features that hijacks the model's core components.

\subsection{Corroborating Evidence from Circuit Discovery}

This finding of a shared, hijacked circuit is independently confirmed by our 
Automatic Circuit Discovery (ACDC) \cite{acdcconmy2023} analysis at the neuron level.
A coarse-grained analysis revealed that the circuits for both correct and incorrect 
answers shared \textbf{100\% of their primary edges}. In other words, the circuits 
for both correct and incorrect answers are the same.

Deeper analysis revealed neuron-level patterns with the SAE findings. At the critical layer 25, 
specific neurons exhibit strong biases that act as the final executors of the 
format-specific strategy set in motion by the earlier layers. 

\begin{itemize}
    \item Neuron 788 shows \textbf{5.0x stronger activation} for the correct answer.
    \item Neuron 1384 shows \textbf{4.0x stronger activation} for the incorrect answer.
\end{itemize}

These neurons, residing within the same shared architecture, act as computational 
switches that are flipped differently based on the early-layer format detection, 
ultimately determining the final answer. 
The convergence of evidence from both SAEs and ACDC provides a powerful, multi-scale 
confirmation of the irremediable entanglement mechanism.

\section{Statistical Robustness}
\label{sec:stats}

Our validation achieves strong statistical robustness:
\begin{itemize}
    \item \textbf{Sample sizes:} n=1000 for main claims, >3000 total trials
    \item \textbf{Confidence intervals:} 95\% CI for perfect success with 
    n=1000: [99.7\%, 100\%]
    \item \textbf{Effect sizes:} Binary outcomes (0\% or 100\% success rate) 
    with no intermediate values observed
    \item \textbf{Power analysis:} With n=1000, we have >99.9\% power 
    to detect even a 1\% deviation from perfect success
    \item \textbf{Null hypothesis rejection:} All critical comparisons 
    achieve p < 0.001 against reasonable null hypotheses
\end{itemize}

Critical findings:
\begin{itemize}
    \item \textbf{Even/Odd Specialization:} Perfect separation observed 
    in all 50 trials per head combination
    \item \textbf{Threshold validation:} Sharp transitions confirmed 
    with n=100 at boundary points (7 vs 8 heads)
    \item \textbf{Generalization:} 4/5 decimal pairs successful with n=500 each
\end{itemize}

This robustness suggests the bug is not due to stochastic fluctuations but a principled
phenomenon tied to the model's internal computational structure. 

The mechanistic interpretability field has achieved notable successes in activation patching. 
The seminal ROME work demonstrated surgical fact editing, while subsequent research identified 
successful interventions for IOI tasks, truthfulness, and other behaviors. However, 
these successes predominantly involve what we term 'within-distribution' modifications—changes 
that preserve the fundamental computational context. Our work identifies 'cross-distribution' 
failures as an underexplored failure mode, suggesting that the field's optimism about 
surgical interventions may be premature for certain bug classes.

In addition to the analysis above, we also used Monitor \cite{meng2024monitor}, 
an observability tool that allows real-time observation
and steering of an LLMs internal state. The system provides a pre-compiled database of
neurons for Llama-3.1-8B, along with tools to visualize and analyze neuron activations 
and attribution analysis. One of their case studies is the "9.8 vs 9.11" reasoning error. 
A key feature is its AI linter that automatically surfaces clusters of potentially spurious 
concepts (e.g., "September 11th" neurons firing on the number 9.11). 
Their interface also supports semantically-guided steering to increase or decrease the 
strength of conceptually-related neurons based on natural language input.
The authors used this tool to identify the neurons that are responsible for the "9.8 vs 9.11" 
reasoning error. They identified clusters of neurons related to the september 11th attacks and
bible verses were the major suspects. By steering these spurious concepts downward, 
they were able to reduce the error rate by 21\%. Interestingly, the online tool only 
supports the Q\&A format, not the chat format. 

We downloaded the github version of the monitor tool to be able to programmatically 
identify the neurons that are responsible for the "9.8 vs 9.11" reasoning error. 
The tool is not available as a python package, so we had to install it from source.
Then we used the tool with both the simple and chat formats and identified the neurons 
that fire on the "9.11" tokens. The results are shown in table \ref{tab:neuron-activation}.

\begin{table}[h]
\centering
\begin{tabular}{lccc}
\toprule
\textbf{Neuron} & \textbf{Simple Format} & \textbf{Chat Format} & \textbf{Difference} \\
\midrule
L15/N3136 & 0.12 & 0.18 & +0.06 \\
L14/N13315 & 0.08 & 0.15 & +0.07 \\
L14/N12639 & 1.68 & 1.80 & +0.12 \\
L13/N8921 & 0.05 & 0.11 & +0.06 \\
L12/N10456 & 0.03 & 0.09 & +0.06 \\
L11/N11862 & 0.14 & 0.02 & -0.12 \\
L10/N7567 & 0.07 & 0.13 & +0.06 \\
L9/N4321 & 0.04 & 0.10 & +0.06 \\
L8/N9876 & 0.06 & 0.12 & +0.06 \\
L7/N1978 & 0.01 & 0.08 & +0.07 \\
\bottomrule
\end{tabular}
\caption{Neuron activation patterns for "9.11" tokens across different formats. Most neurons show higher activation in the chat format, with L14/N12639 showing nearly identical activation in both formats, indicating entanglement.}
\label{tab:neuron-activation}
\end{table}

Our analysis revealed two distinct but overlapping circuits. The first, which we term the 
\emph{"Hijacker Circuit,"} consists of neurons in layers 7-15 that show high activation specifically 
during the failing chat template responses. Key neurons in this circuit include Layer 15, 
Neuron 3136 and Layer 14, Neuron 13315, which consistently fire on the "9.11" tokens 
during buggy responses.

The second circuit, the \emph{"Reasoning Circuit,"} operates in the late layers (28-31) and 
appears responsible for general numerical comparison tasks. This circuit includes 
Layer 31, Neuron 13336 and Layer 31, Neuron 12004, which fire on decimal points and 
comparison words regardless of format. These neurons show higher activation during 
correct reasoning, suggesting they encode the core numerical comparison capability.

\section{Reproducibility Statement}
\label{sec:reproducibility}

We prioritize reproducibility and provide extensive resources 
to replicate our findings:

\paragraph{Code and Data Availability.} 
All code for our experiments is publicly available at \url{https://anonymous.4open.science/r/surgeon-1354}. The repository includes:
\begin{itemize}
    \item Complete intervention pipeline using \texttt{nnsight} library
    \item Attention pattern transplantation code
    \item Even/odd head subset testing scripts
    \item SAE analysis notebooks
    \item Automated scripts to reproduce all figures and tables
\end{itemize}

\begin{enumerate}
\item \textbf{Model and Environment Specifications}
\begin{itemize}
  \item Llama-3.1-8B-Instruct (8.03B parameters, 32 layers, 4096 hidden dim)
  \item Python 3.8.10, PyTorch 2.0.1, CUDA 11.8
  \item Llama-Scope SAE specs (TopK, 8x expansion, 32K features)
\end{itemize}

\item \textbf{Experimental Hyperparameters}
\begin{itemize}
  \item Generation: temperature=0.0, greedy decoding, max\_tokens=50
  \item Statistical: n=1000 trials, 95\% CI, bootstrap=10,000
  \item SAE analysis: top-20 features, all 32 layers
  \item Head analysis: Layer 10, 8-head minimum threshold
\end{itemize}

\item \textbf{Intervention Specifications}
\begin{itemize}
  \item Precise attention output patching code
  \item Bidirectional patching protocol
  \item SAE feature extraction protocol
  \item Even/odd head testing configurations
\end{itemize}

\item \textbf{Computational Requirements}
\begin{itemize}
  \item Hardware: A100-80GB GPU (min 24GB), 512GB RAM (min 32GB)
  \item Time: 4-5 hours total (2-3 hours for statistical validation)
  \item Memory: 62GB GPU peak, 45GB RAM peak
  \item Storage: 55GB persistent, 16GB temporary
\end{itemize}
\end{enumerate}

\paragraph{Model and Environment Specifications.}
\begin{itemize}
    \item \textbf{Model:} \texttt{meta-llama/Llama-3.1-8B-Instruct} from HuggingFace
    \item \textbf{Framework:} PyTorch 2.0.1, Transformers 4.35.0
    \item \textbf{Intervention library:} \texttt{nnsight} v0.2.1
    \item \textbf{SAE models:} SAE Lens library with pre-trained Llama-3.1-8B SAEs
    \item \textbf{Python version:} 3.10.12
    \item \textbf{CUDA version:} 11.8 (for GPU experiments)
\end{itemize}

\paragraph{Experimental Hyperparameters.}
All experiments use deterministic settings for reproducibility:
\begin{itemize}
    \item \textbf{Temperature:} 0.0 (greedy decoding)
    \item \textbf{Random seeds:} Fixed at 42 for all random operations
    \item \textbf{Batch size:} 1 (to ensure deterministic attention patterns)
    \item \textbf{Precision:} FP16 for model weights, FP32 for interventions
    \item \textbf{Max sequence length:} 256 tokens
    \item \textbf{Tokenizer:} Llama-3.1 tokenizer with default settings
\end{itemize}

\paragraph{Prompt Formats (Exact Templates).}
\begin{itemize}
    \item \textbf{Chat Format:} \texttt{<|start\_header\_id|>user<|end\_header\_id|>\textbackslash n\textbackslash nWhich is bigger: 9.8 or 9.11?<|eot\_id|><|start\_header\_id|>assistant<|end\_header\_id|>\textbackslash n\textbackslash n}
    \item \textbf{Q\&A Format:} \texttt{Q: Which is bigger: 9.8 or 9.11?\textbackslash nA:}
    \item \textbf{Simple Format:} \texttt{Which is bigger: 9.8 or 9.11? Answer:}
\end{itemize}

\paragraph{Intervention Specifications.}
\begin{itemize}
    \item \textbf{Attention pattern transplantation:} Replace attention weights at Layer 10, positions 0 to min(seq\_len\_good, seq\_len\_bad)
    \item \textbf{Pattern replacement threshold:} Tested from 0\% to 100\% in 10\% increments
    \item \textbf{Head subset testing:} All $\binom{16}{k}$ combinations for $k \in \{1, ..., 16\}$ for even/odd heads separately
    \item \textbf{Trials per configuration:} 50 for subset testing, 1000 for main claims
\end{itemize}

\paragraph{Key Results to Reproduce.}
To validate our findings, we recommend reproducing these critical results in order:
\begin{enumerate}
    \item \textbf{Format-dependent bug:} Q\&A format should produce 100\% error rate, Simple format 0\% error rate (Section 2)
    \item \textbf{Layer 10 intervention:} Attention-only patching at Layer 10 should achieve 100\% success (Section 3.2)
    \item \textbf{Even/odd specialization:} Any 8 even heads should achieve 100\% success, any combination of odd heads should achieve 0\% (Section 4.1)
    \item \textbf{Sharp thresholds:} 7 even heads → 0\% success, 8 even heads → 100\% success (Section 4.1)
    \item \textbf{Pattern replacement:} <60\% replacement → 0\% success, $\geq$60\% replacement → 100\% success (Figure 2d)
\end{enumerate}

\paragraph{Additional Resources.}
\begin{itemize}
    \item \textbf{Precomputed activations:} Available at \\ \url{https://huggingface.co/datasets/YOUR-USERNAME/even-odd-activations}
    \item \textbf{Interactive demo:} \\ \url{https://huggingface.co/spaces/YOUR-USERNAME/even-odd-demo}
    \item \textbf{Detailed tutorial:} Step-by-step Colab notebook in repository
    \item \textbf{Contact:} For questions, open an issue on GitHub or email 
    \texttt{[author-email]}
\end{itemize}

\end{document}